\documentclass[10pt,twocolumn]{article}

\usepackage[letterpaper,left=0.75in,right=0.75in,top=1in,bottom=1in]{geometry}

\usepackage{microtype}
\usepackage{graphicx}
\usepackage{subfigure}
\usepackage{booktabs}

\usepackage[table]{xcolor}
\definecolor{lightgreen}{RGB}{220,245,220}

\usepackage{hyperref}

\usepackage{algorithm}
\usepackage{algorithmic}

\usepackage{amsmath}
\usepackage{amssymb}
\usepackage{mathtools}
\usepackage{amsthm}

\usepackage[round]{natbib}
\usepackage[capitalize,noabbrev]{cleveref}

\usepackage[textsize=tiny]{todonotes}

\usepackage{authblk}

\title{A Semi-Supervised Pipeline for Generalized Behavior Discovery from Animal-Borne Motion Time Series}

\author[1]{Fatemeh Karimi Nejadasl}
\author[1]{Judy Shamoun-Baranes}
\author[1]{Eldar Rakhimberdiev}

\affil[1]{Department of Theoretical and Computational Ecology, Institute for Biodiversity and Ecosystem Dynamics, University of Amsterdam, Amsterdam, The Netherlands}
\affil[ ]{\texttt{f.kariminejadasl@uva.nl, j.z.shamoun-baranes@uva.nl, e.n.rakhimberdiev@uva.nl}}

\date{}

\begin{document}
\maketitle

\begin{abstract}
Learning behavioral taxonomies from animal-borne sensors is challenging because labels are scarce, classes are highly imbalanced, and behaviors may be absent from the annotated set. We study generalized behavior discovery in short multivariate motion snippets from gulls, where each sample is a sequence with 3-axis IMU acceleration (20 Hz) and GPS speed, spanning nine expert-annotated behavior categories. We propose a semi-supervised discovery pipeline that (i) learns an embedding function from the labeled subset, (ii) performs label-guided clustering over embeddings of both labeled and unlabeled samples to form candidate behavior groups, and (iii) decides whether a discovered group is truly novel using a containment score. Our key contribution is a KDE + HDR (highest-density region) containment score that measures how much a discovered cluster distribution is contained within, or contains, each known-class distribution; the best-match containment score serves as an interpretable novelty statistic. In experiments where an entire behavior is withheld from supervision and appears only in the unlabeled pool, the method recovers a distinct cluster and the containment score flags novelty via low overlap, while a negative-control setting with no novel behavior yields consistently higher overlaps. These results suggest that HDR-based containment provides a practical, quantitative test for generalized class discovery in ecological motion time series under limited annotation and severe class imbalance.
\end{abstract}

\begin{figure}[t]
  \centering
   \includegraphics[width=0.98\linewidth]{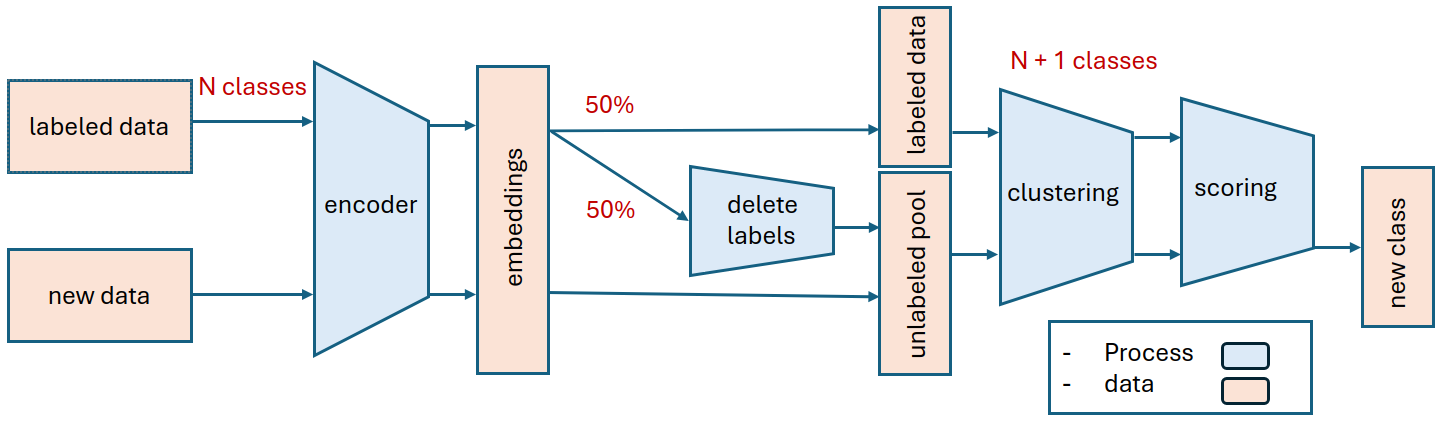}
   \caption{\textbf{Generalized behavior discovery from animal-borne motion.}
(1) \textbf{Supervised representation learning:} train encoder $f_\theta$ on labeled data $\mathcal{D}_L$ to embed IMU+GPS snippets (\cref{sec:rep_learning}).
(2) \textbf{Semi-supervised clustering:} embed $\mathcal{D}_L \cup \mathcal{D}_U$ and run label-guided K-means to form candidate behaviors (\cref{sec:ss-kmeans-method}).
(3) \textbf{Novelty decision:} compute a KDE+HDR containment score vs. known classes and flag clusters with low best-match score $O_c$ as novel (\cref{sec:containment_score}).
\textbf{Discovery:} withhold one class or use only known behaviors in $\mathcal{D}_U$ (negative control)(\cref{sec:protocols}).
\textbf{Deployment:} apply the same pipeline to sliding windows of unlabeled data (\cref{sec:deployment}).}
   \label{fig:pipeline}
\end{figure}

\section{Introduction}
\label{sec:introduction}
Discovering general behavioral patterns from animal activity data remains a challenge in movement ecology and behavioral science. In many ecological studies, near-continuous motion data, such as those collected from inertial sensors and GPS trackers contain latent behavioral states that are difficult or even impossible to annotate exhaustively. The expert annotator must be physically present at the location and during the moment of occurrence of the behavior of interest and in many cases these places are inaccessible (e.g. behavior at sea) or unobservable (e.g. behavior at night, or behavior exercised only in absence of human observer). Consequently, there is a need for semi-supervised frameworks capable of both recognizing known behaviors and identifying new, previously unobserved ones.

In this work, we introduce a pipeline for general behavior discovery in bird movement data, focusing on a dataset collected from gulls equipped with inertial measurement units (IMUs) and GPS \cite{camphuysen2015sexually,van2019foraging}. The solar-powered tracking devices were attached on the back of the animal \citep{bouten2013flexible}. Each sample comprises a short temporal sequence (20 time steps) with four channels: three axes of acceleration and GPS speed. The dataset includes nine, expert annotated behavioral classes \citep{shamoun2016flap} but suffers from limited annotation and imbalance across classes.

We study generalized behavior discovery in a setting that reflects how animal-borne sensor data are used in practice. First, an ecologist typically has access to a small labeled subset of behaviors from focal observations or curated annotations, while the bulk of the sensor stream remains unlabeled. Second, the unlabeled stream can contain familiar behaviors with genuinely novel ones. This motivates a \emph{generalized category discovery} (GCD) formulation in which an unlabeled pool contains a mixture of known and unknown classes, and the goal is to recover coherent clusters while deciding whether any cluster is plausibly novel.

To address this, we propose a semi-supervised discovery pipeline with two complementary uses. In the \emph{discovery protocols}, we validate the pipeline and calibrate a novelty operating point by constructing unlabeled pools from annotated data in a controlled way. Specifically, we (i) withhold an entire behavior type from labeled data and place it only in the unlabeled pool to test whether the pipeline can recover it as a distinct cluster, and (ii) run a negative-control protocol where the unlabeled pool contains only known behaviors, so any extra cluster should be explainable by known-class structure. In \emph{deployment}, we apply the same procedure to sliding windows from unlabeled streams, where the unlabeled pool naturally contains a mixture of known behaviors and candidate novel segments. Across both settings, the algorithmic steps are identical; only the composition of the unlabeled pool changes.

Our pipeline (\cref{fig:pipeline}) combines label-guided clustering with an explicit novelty decision statistic. We first learn an embedding function from the labeled subset. We then perform semi-supervised K-means on embeddings of labeled and unlabeled samples, allocating an additional free cluster to absorb candidate novel structure. Finally, we decide whether a discovered cluster is plausibly novel by comparing its embedding distribution to each known-class distribution using a KDE + highest-density-region (HDR) containment score. Intuitively, the containment score measures whether the mass of a discovered cluster lies within the high-density region of a known class (or vice versa). The best-match containment score provides an interpretable novelty statistic that supports a simple operating rule: low containment suggests novelty, while high containment suggests re-partitioning of known behaviors.

Empirically, when an entire behavior is withheld from supervision and appears only in the unlabeled pool, the method typically recovers a distinct cluster and the containment score flags novelty via low overlap. In contrast, under the negative-control protocol with no novel behavior, containment scores remain consistently higher, indicating that the extra cluster is not genuinely novel. These results suggest that HDR-based containment offers a practical, quantitative test for generalized behavior discovery in ecological motion time series under limited annotation and severe class imbalance.

The contributions of this paper are:
\begin{itemize}
    \item \textbf{Task and data.} Semi-supervised generalized behavior discovery from short multivariate motion snippets (3-axis IMU acceleration + GPS speed) with limited labels and strong class imbalance across nine expert-annotated behaviors.
    \item \textbf{Method.} A label-guided semi-supervised K-means pipeline that jointly clusters labeled and unlabeled embeddings and allocates an additional free cluster for candidate novel structure.
    \item \textbf{Novelty decision.} A KDE + HDR containment score that compares discovered clusters to known-class embedding distributions and yields an interpretable cluster-level novelty statistic, validated via withheld-class and negative-control protocols and reused unchanged for stream deployment.
\end{itemize}

\section{Related Work}
\label{sec:related_work}
\textbf{Behavior classification from animal-borne sensors.}
Animal-borne accelerometers and GPS enable fine-scale inference of activity and behavior when direct observation is limited \citep{kays2015terrestrial}. A large body of work treats behavior recognition as a supervised learning problem, using labeled segments from focal observations or video to train classifiers on tri-axial acceleration features and related signals \citep{nathan2012using,bidder2014love,resheff2014accelerater,shamoun2016flap}. These pipelines have become common in movement ecology and biologging, but they typically assume a \emph{closed} label set: new data are forced into the predefined classes even when novel behaviors are present. Recent work has highlighted this as an open-set recognition failure mode for accelerometer-based behavior classification \citep{wilson2026ignoring}. Our setting is motivated by the same practical limitation, but we focus on \emph{discovering} candidate novel behaviors rather than only rejecting them as unknown.

\textbf{Unsupervised and weakly supervised behavior discovery in ecology.}
Beyond discriminative classifiers, ecologists have long used unsupervised and latent-state approaches to infer behavioral modes from sequential sensor streams. For example, ethograms can be generated by transforming accelerometer signals and applying unsupervised clustering to discover recurring motion patterns \citep{sakamoto2009can}. In particular, hidden Markov models (HMMs) provide a principled way to model serial dependence and recover activity states from accelerometer sequences, either in supervised form (state labels available) or in unsupervised form (states discovered as latent modes) \citep{leos2017analysis}. While HMMs are attractive for sequential structure, the resulting latent states may not align with ecologically meaningful behaviors, and deciding whether a discovered state corresponds to a genuinely new category is often qualitative. Our pipeline is complementary: we learn an embedding suited to short motion snippets and use label-guided clustering to propose groups, then apply an explicit quantitative novelty test based on distributional overlap.

\textbf{Novel class discovery, generalized category discovery, and open-world recognition.}
In machine learning, \emph{novel class discovery} (NCD) studies how to cluster unlabeled examples from unseen classes while leveraging labeled data from different known classes \citep{han2019learning,fini2021unified,han2021autonovel}. \emph{Generalized category discovery} (GCD) relaxes NCD by allowing the unlabeled pool to contain a mixture of known and unknown classes, and proposes semi-supervised clustering objectives to separate and label both \citep{vaze2022gcd}. This formulation is a close match to animal-borne sensing studies in movement ecology, where unlabeled streams usually include both familiar and unfamiliar behaviors. However, existing GCD and NCD work is dominated by image benchmarks, and novelty is commonly assessed via cluster purity, heuristics, or downstream labeling effort. We adapt the GCD perspective to multivariate motion time series with severe imbalance and very short measurement windows, and we introduce a decision statistic tailored to the question ecologists face: whether a discovered cluster is plausibly a new behavior rather than a fragment or re-partitioning of known ones.

\textbf{Representation learning for time series.}
Self-supervised time-series representation learning has advanced rapidly, including contrastive and neighborhood-based objectives such as TS2Vec and temporal neighborhood coding \citep{tonekaboni2021unsupervised,yue2022ts2vec}. These methods can reduce label dependence, but in short-window biologging settings, there is limited temporal context to construct multiple informative views of each example, and class imbalance can create many false negatives when using instance-discrimination losses. Motivated by these constraints, we use a lightweight supervised embedding model, trained on the labeled subset, and delegate novelty detection to a distributional comparison stage rather than relying on representation learning alone.

\textbf{Novelty decisions via distributional overlap and HDRs.}
A core challenge in behavior discovery is turning clusters into actionable scientific hypotheses. Instead of relying on visual inspection or ad hoc thresholds, we quantify similarity between a discovered cluster and each known class using overlap of highest density regions (HDRs) estimated from KDE \citep{hyndman1996computing,SamworthWand2010}. This connects to classical overlap coefficients for distribution agreement \citep{inman1989overlapping,schmid2006nonparametric}, but yields an interpretable novelty statistic at the cluster level: low best-match containment indicates that the cluster occupies a region of embedding space not well explained by any labeled behavior.

\begin{table}[t]
\centering
\caption{Existing novel class discovery. For each withheld behavior (rem class), we report the matched discovered cluster (disc class), withheld-class accuracy (acc), and the containment score (cnt score) (best-match $O_c$). Lower containment indicates a cluster that is poorly explained by any known class.}
\label{tab:existing}
\begin{tabular}{l r r r r} 
\hline


ind:name & rem class& disc class& acc& cnt score\\
0:Flap     & 0 & 0 & 0.933 & 0.144 \\
1:ExFlap   & 1 & 1 & 0.789 & 0.101 \\
2:Soar     & 2 & 2 & 0.279 & 0.019 \\
3:Boat     & 3 & 3 & 0.591 & 0.158 \\
4:Float    & 4 & 4 & 0.554 & 0.059 \\
5:SitStand & 5 & 5 & 0.798 & 0.028 \\
6:TerLoco  & 6 & 6 & 0.332 & 0.153 \\
\rowcolor{lightgreen} 8:Manouvre & 8 & 8 & 0.000     & 0.496 \\
9:Pecking  & 9 & 9 & 0.582 & 0.063 \\
\hline
\end{tabular}

\vspace{1em}

\caption{Non-existing class discovery (negative control). The pipeline still allocates an extra cluster (discover class 10), but containment scores remain above the novelty threshold $0.3$, indicating that the extra cluster is not genuinely novel.}
\label{tab:nonexisting}
\begin{tabular}{l r r r r}
\hline
ind:name & rem class & disc class & acc & cnt score  \\
\hline

0:Flap     & 0 & 10 & - & 0.674 \\
1:ExFlap   & 1 & 10 & - & 0.605 \\
2:Soar     & 2 & 10 & - & 0.384 \\
3:Boat     & 3 & 10 & - & 0.516 \\
4:Float    & 4 & 10 & - & 0.562 \\
5:SitStand & 5 & 10 & - & 0.529 \\
6:TerLoco  & 6 & 10 & - & 0.417 \\
8:Manouvre & 8 & 10 & - & 0.471 \\
9:Pecking  & 9 & 10 & - & 0.472 \\
\hline
\end{tabular}
\end{table}

\begin{figure}[t]
  \centering
   \includegraphics[width=0.98\linewidth]{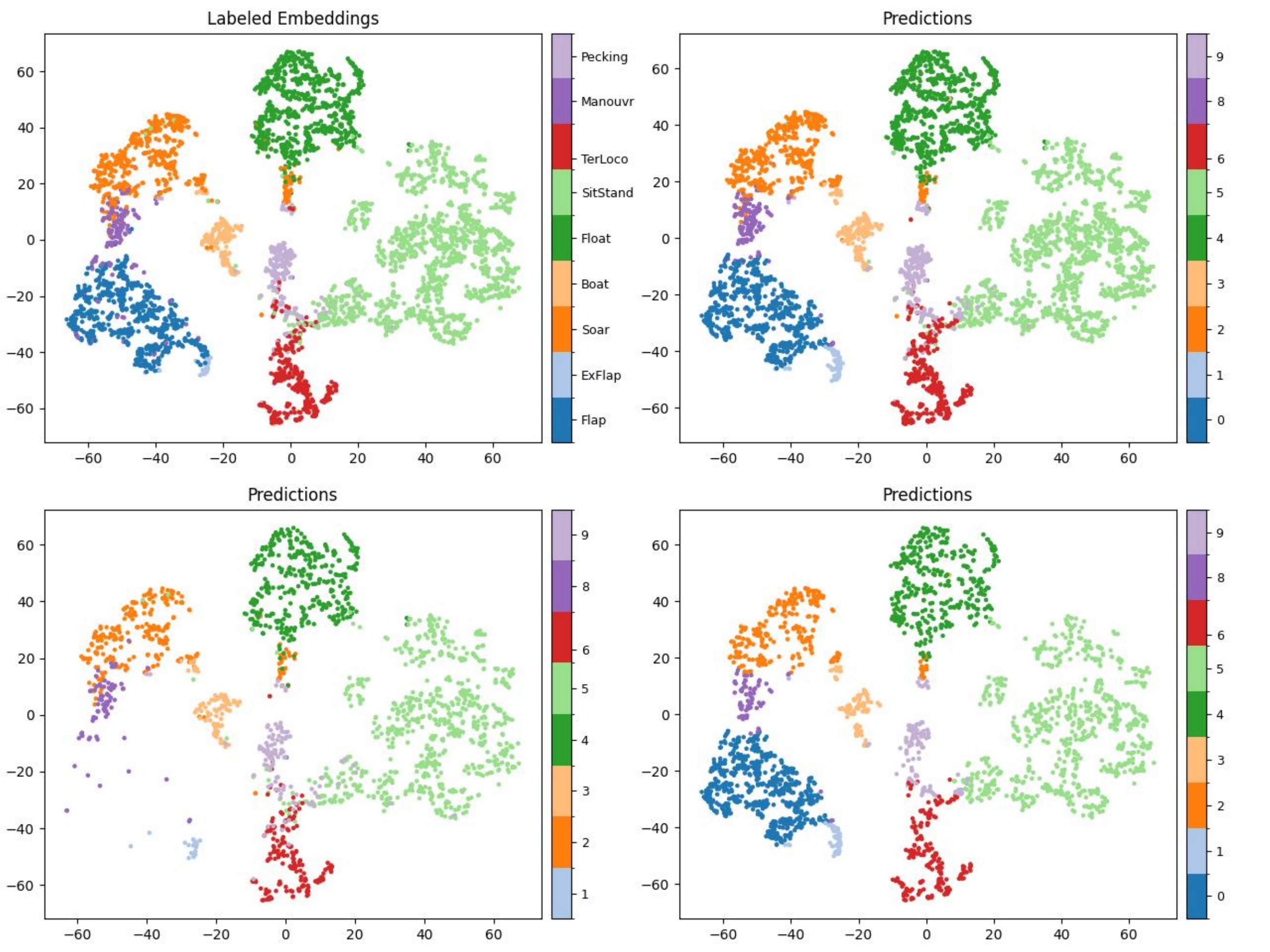}
   \caption{\textbf{Representative existing-novel discovery run (withheld Flap).} t-SNE of embeddings for \texttt{0:Flap}. \textbf{Top left:} ground-truth labels. \textbf{Top right:} cluster assignments (predictions) for all data $\mathcal{D}_L \cup \mathcal{D}_U$. \textbf{Bottom left:} assignments restricted to labeled data $\mathcal{D}_L$. \textbf{Bottom right:} assignments restricted to unlabeled data $\mathcal{D}_U$. Axes show t-SNE coordinates (arbitrary units).}

   \label{fig:tsne_d0}
\end{figure}

\begin{figure}[t]
  \centering
   \includegraphics[width=0.98\linewidth]{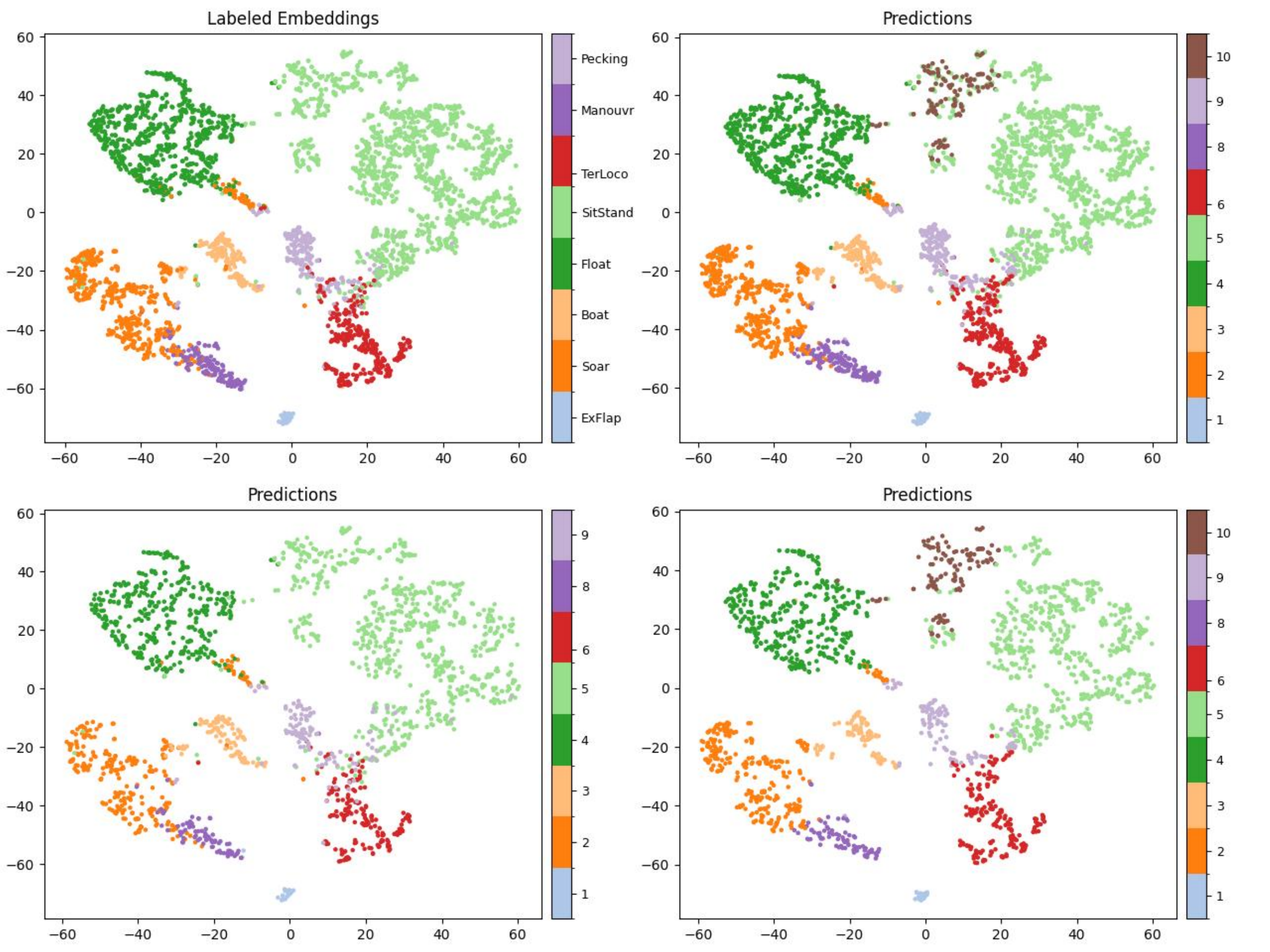}
   \caption{\textbf{Representative negative-control run (no novel class; removed Flap for the trial).} t-SNE of embeddings with an extra cluster allocated. \textbf{Top left:} ground-truth labels. \textbf{Top right:} cluster assignments (predictions) for all data $\mathcal{D}_L \cup \mathcal{D}_U$. \textbf{Bottom left:} assignments restricted to labeled data $\mathcal{D}_L$. \textbf{Bottom right:} assignments restricted to unlabeled data $\mathcal{D}_U$. Axes show t-SNE coordinates (arbitrary units).}

   \label{fig:tsne_r0_d10}
\end{figure}

\begin{figure*}[t]
  \centering
   \includegraphics[width=0.98\linewidth]{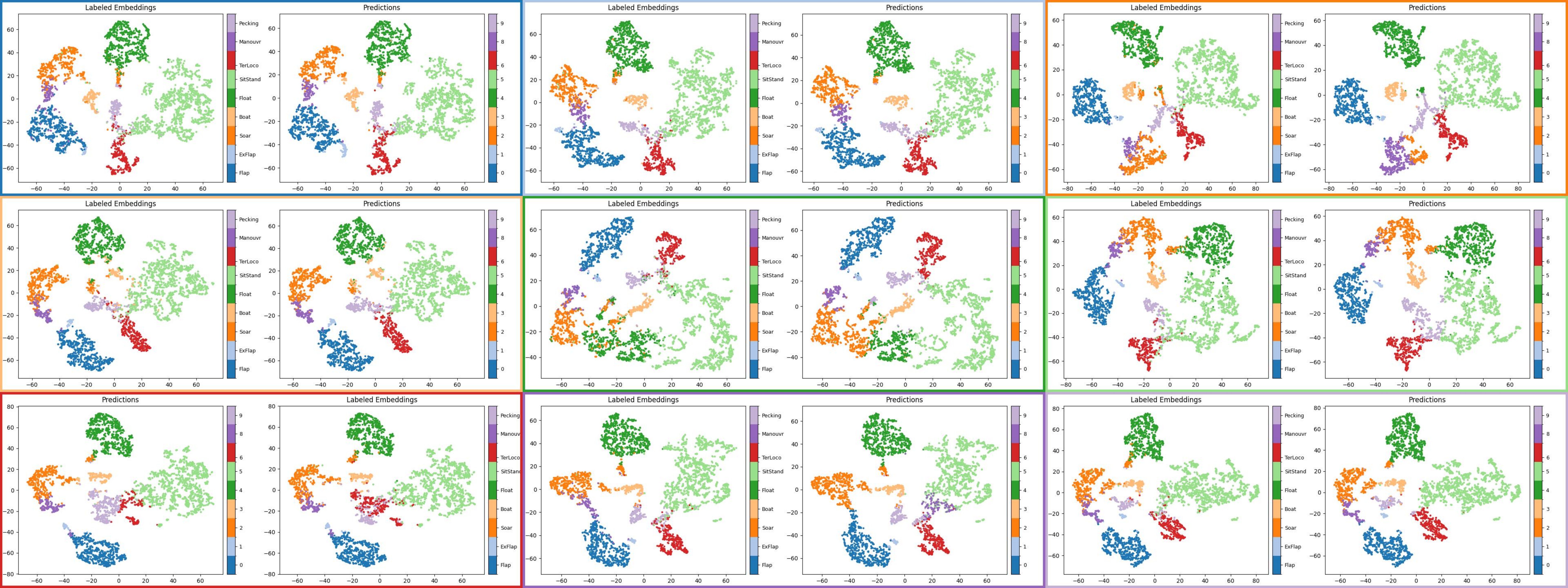}
    \caption{\textbf{Existing novel class discovery across all withheld behaviors.}
    For each withheld behavior, we show a pair of t-SNE plots: \textbf{ground-truth labels} (left) and \textbf{cluster assignments for all data} $\mathcal{D}_L \cup \mathcal{D}_U$ (right). The nine withheld-behavior pairs are arranged from left to right and top to bottom in the following order: 0:Flap, 1:ExFlap, 2:Soar, 3:Boat, 4:Float, 5:SitStand, 6:TerLoco, 8:Manouvre, 9:Pecking. The border color of each pair indicates the removed class for that trial. Point colors in the left plots indicate ground-truth classes; point colors in the right plots indicate predicted cluster IDs. Axes show t-SNE coordinates (arbitrary units).}
   \label{fig:all_tsne_lp}
\end{figure*}

\begin{figure*}[t]
  \centering
   \includegraphics[width=0.98\linewidth]{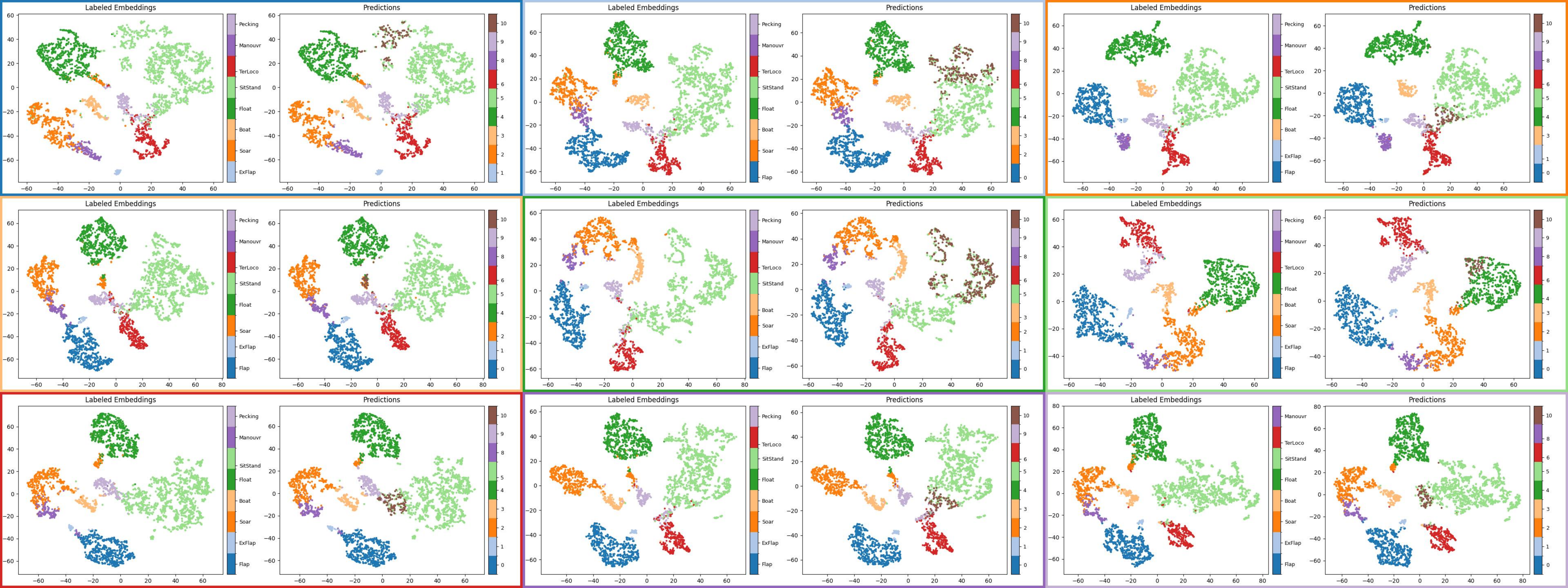}
    \caption{\textbf{Negative-control (non-existing) class discovery across trials.}
    For each removed-class trial, we show a pair of t-SNE plots: \textbf{ground-truth labels} (left) and \textbf{cluster assignments for all data} $\mathcal{D}_L \cup \mathcal{D}_U$ (right), while still allocating one extra cluster. The nine trial pairs are arranged from left to right and top to bottom in the following order: 0:Flap, 1:ExFlap, 2:Soar, 3:Boat, 4:Float, 5:SitStand, 6:TerLoco, 8:Manouvre, 9:Pecking. The border color of each pair indicates the removed class for that trial. Point colors in the left plots indicate ground-truth classes; point colors in the right plots indicate predicted cluster IDs. Axes show t-SNE coordinates (arbitrary units).}
   \label{fig:all_tsne_lp10}
\end{figure*}

\section{Methodology}
\label{sec:methodology}

\subsection{Overview}
We address generalized category discovery (GCD) for short motion snippets under limited annotation. The pipeline has three stages (\cref{fig:pipeline}): (1) learn an embedding function $f_\theta$ from labeled data, (2) run label-guided semi-supervised K-means on embeddings of labeled and unlabeled samples, and (3) decide whether any discovered cluster is plausibly novel via a distributional containment statistic.

We use the same pipeline in two roles. First, we run controlled \emph{discovery protocols} (\cref{sec:protocols}) where the unlabeled pool is constructed from labeled data to (i) test whether a withheld behavior can be recovered and (ii) calibrate an operating threshold for the containment score. Second, we apply the identical procedure to \emph{unlabeled streams} in deployment, where the unlabeled pool is formed from sliding windows of stream segments with labels removed. The algorithmic steps remain unchanged; only the source and composition of unlabeled data differ.

\subsection{Data Representation}
Each motion sequence is represented as a multivariate time series of length 20 (1 second of measurements with 20 Hz frequency) with four input channels: three axes of acceleration from the IMU sensor and one GPS-derived speed signal. The dataset consists of 4,338 samples (\cref{tab:behavior-counts}) categorized into nine behavioral classes. The last channel, GPS speed is divided by 22 $m/s$ and the other accelerometer channels are cropped between -2 and 2 g. Temporal augmentations such as random jitter, scaling, time and magnitude warping were tried during training but did not improve model robustness. Therefore, the model was trained without data augmentation.

\begin{table}[t]
\caption{Behavior counts in the gull dataset.}
\label{tab:behavior-counts}
\vskip 0.15in
\centering
\begin{tabular}{lr}
\toprule
\textbf{Behavior} & \textbf{Count} \\
\midrule
Flap      & 643 \\
ExFlap    & 38 \\
Soar      & 537 \\
Boat      & 176 \\
Float     & 729 \\
SitStand  & 1502 \\
TerLoco   & 337 \\
Manouvre  & 151 \\
Pecking   & 225 \\
\bottomrule
\end{tabular}
\vskip -0.1in
\end{table}

\subsection{Representation Learning}
\label{sec:rep_learning}
We learn an embedding function $f_\theta$ from the labeled subset $\mathcal{D}_L$ using a standard supervised objective. Concretely, we train a lightweight classifier and use its pre-softmax outputs (logits) as the representation $z=f_\theta(x)$ for subsequent clustering and distributional comparison. This choice yields a task-aligned embedding while keeping the discovery pipeline encoder-agnostic: any supervised or self-supervised time-series encoder could be substituted for $f_\theta$ without changing the later stages.

\subsection{Semi-Supervised Clustering Setup}
\label{sec:ss-kmeans-method}
We adopt a label-guided semi-supervised K-means procedure in the spirit of generalized category discovery \citep{vaze2022gcd}. We assume the data are partitioned into a labeled subset and an unlabeled subset,
\begin{align}
\mathcal{D}_L &= \{(x_i, y_i)\}_{i=1}^{N_L}, \\
\mathcal{D}_U &= \{x_j\}_{j=1}^{N_U},
\end{align}
where $\mathcal{D}_L$ contains samples from the set of known behaviors and $\mathcal{D}_U$ may contain a mixture of known and previously unseen behaviors.

Given embeddings $z=f_\theta(x)$ for all samples in $\mathcal{D}_L \cup \mathcal{D}_U$, we run semi-supervised K-means jointly over labeled and unlabeled embeddings. The labels in $\mathcal{D}_L$ guide the clustering so that clusters corresponding to known behaviors are encouraged to be consistent with the labeled classes, while the assignments for samples in $\mathcal{D}_U$ are inferred. We allocate one additional free cluster beyond the number of known classes to capture structure not well explained by the labeled classes, yielding a candidate novel group when appropriate.

The concrete construction of $\mathcal{D}_L$ and $\mathcal{D}_U$ for controlled discovery protocols and deployment on unlabeled streams is given in \cref{sec:dis_dep}.

\subsection{HDR-Containment Score via KDE}
\label{sec:containment_score}
To quantify novelty, we compare the discovered cluster density \(p_c(z)\) to each labeled class density \(p_k(z)\) using containment scores derived from \(\alpha\)-high density regions (HDRs) \citep{hyndman1996computing,SamworthWand2010}. We estimate these densities in a 2D t-SNE projection of \(z\) for tractable KDE and HDR computation, so the score is a low-dimensional overlap heuristic rather than an exact embedding-space density overlap measure. HDRs are density level sets with fixed probability content, estimated via Gaussian kernel density estimators. The measure is inspired by classical overlap coefficients \citep{inman1989overlapping,schmid2006nonparametric}, but operates on \(\alpha\)-HDRs rather than full densities.

\textbf{HDRs from KDE.}
We fit Gaussian kernel density estimators (KDEs) to obtain \(\hat{p}_c\) and \(\hat{p}_k\). For a fixed \(\alpha \in (0,1)\), the \(\alpha\)-HDR of a density \(p\) is
\[
S_p(\alpha) \;=\; \{\, z : p(z) \ge t_p(\alpha) \,\}, \quad \text{with } \int_{S_p(\alpha)} p(z)\,\mathrm{d}z = \alpha .
\]
In practice, \(t_p(\alpha)\) is estimated by sampling from the KDE and taking the \((1-\alpha)\)-quantile of the sampled log-densities.

\textbf{Directional containment probabilities.}
For a pair \((c,k)\), define the probability under one distribution of falling inside the other's HDR\footnote{We approximate these probabilities by Monte Carlo sampling from the fitted KDEs in the 2D projection.}:
\[
\begin{aligned}
P_{c\to k}(\alpha) &= \int \mathbf{1}\!\left\{ z \in S_{p_k}(\alpha) \right\}\, p_c(z)\,\mathrm{d}z,\\
P_{k\to c}(\alpha) &= \int \mathbf{1}\!\left\{ z \in S_{p_c}(\alpha) \right\}\, p_k(z)\,\mathrm{d}z.
\end{aligned}
\]

We normalize by \(\alpha\) (since each HDR has mass \(\alpha\) under its own density) and cap at 1:
\[
\begin{aligned}
C_{c\to k}(\alpha) &= \min\!\left(1,\frac{P_{c\to k}(\alpha)}{\alpha}\right),\\
C_{k\to c}(\alpha) &= \min\!\left(1,\frac{P_{k\to c}(\alpha)}{\alpha}\right).
\end{aligned}
\]

\textbf{Symmetric containment score.}
Our final containment score is the maximum of the two directions:
\[
\operatorname{Contain}(c,k;\alpha) \;=\; \max\big\{\, C_{c\to k}(\alpha),\; C_{k\to c}(\alpha) \,\big\}.
\]
This quantity lies in \([0,1]\); it equals 1 when one \(\alpha\)-HDR is contained in the other, and approaches 0 when the HDRs are disjoint.

\textbf{Cluster-level summary and novelty decision.}
For each discovered cluster \(c\), we report its best match among known classes:
\[
O_c \;=\; \max_k \; \operatorname{Contain}(c,k;\alpha).
\]
We label \(c\) as a novel class if \(O_c < 0.3\); otherwise, \(c\) is considered to correspond to an existing behavioral category.

\subsection{Discovery vs Deployment}
\label{sec:dis_dep}

\subsubsection{Discovery Protocols and Evaluation}
\label{sec:protocols}
We describe two controlled discovery protocols that reuse labeled data to simulate unlabeled pools. These protocols do not alter the pipeline; they validate discovery behavior and provide a practical way to calibrate an operating point for $O_c$ that is later used for stream analysis.

We consider two settings: (i) \emph{existing novel class discovery}, where the unlabeled pool contains a truly unseen behavior, and (ii) \emph{non-existing class discovery} (negative control), where the unlabeled pool contains only known behaviors. In both cases, the same pipeline is applied; only the construction of $\mathcal{D}_U$ changes.

\textbf{Data splits.}
Let $\mathcal{Y}$ denote the set of annotated behaviors. For each trial, we designate one behavior $y^\star \in \mathcal{Y}$ as the \emph{withheld} behavior. For all $y \in \mathcal{Y}\setminus\{y^\star\}$, we split samples of class $y$ into two disjoint halves: a labeled subset $\mathcal{D}_L$ (labels retained) and an unlabeled subset $\mathcal{D}_U^{\text{known}}$ (labels available for evaluation but not used by the method).

\textbf{Existing novel class discovery.}
To simulate a genuinely unseen behavior, we add all samples of $y^\star$ to the unlabeled pool:
\[
\mathcal{D}_U \;=\; \mathcal{D}_U^{\text{known}} \;\cup\; \mathcal{D}_U^{\text{novel}},
\quad \text{where } \mathcal{D}_U^{\text{novel}}=\{x : y=y^\star\}.
\]
We train the encoder $f_\theta$ on $\mathcal{D}_L$ only (no access to $y^\star$), then run semi-supervised K-means on embeddings from $\mathcal{D}_L \cup \mathcal{D}_U$. We report withheld-class accuracy and the containment score $O_c$ for the discovered cluster matched to the withheld behavior.

\textbf{Non-existing class discovery (negative control).}
To test whether the pipeline spuriously flags novelty when none exists, we keep all behaviors in the labeled set and construct the unlabeled pool using only known-class samples:
\[
\mathcal{D}_U \;=\; \mathcal{D}_U^{\text{known}} \quad (\text{no } \mathcal{D}_U^{\text{novel}}).
\]
We run the same semi-supervised K-means procedure while still allocating one additional free cluster beyond the known classes. In this setting, the extra cluster should correspond to a re-partitioning of known behaviors and thus attain a high best-match containment score. We use this protocol to calibrate a novelty operating point and, in our experiments, flag novelty when $O_c < 0.3$.

\subsubsection{Deployment Protocol on Unlabeled Streams}
\label{sec:deployment}
For deployment, we split the available labeled data 50/50 into $\mathcal{D}_L$ and $\mathcal{D}_U^{\text{known}}$ by removing labels from the latter. We train the encoder on $\mathcal{D}_L$. The unlabeled pool includes (i) $\mathcal{D}_U^{\text{known}}$ and (ii) a sliding window of unlabeled stream segments, denoted $\mathcal{D}_U^{\text{novel}}$:
\[
\mathcal{D}_U \;=\; \mathcal{D}_U^{\text{known}} \;\cup\; \mathcal{D}_U^{\text{novel}}.
\]
We then run semi-supervised K-means with one additional free cluster on $\mathcal{D}_L \cup \mathcal{D}_U$ and compute $O_c$ for the newly formed clusters to decide whether a candidate behavior is plausibly novel.

\subsection{Implementation Details}
The feature extractor $f_\theta$ is a lightweight temporal convolutional network with three one-dimensional convolutional layers. Each layer is followed by batch normalization, ReLU, and dropout with rate 0.25. All convolutions use kernel size 3 with padding 1, and each layer outputs 30 channels.

The input consists of four channels: three-axis IMU linear acceleration and GPS speed. Global average pooling over time after the final convolutional block yields a fixed-length representation. We use the classifier’s pre-softmax logits as embeddings for clustering.

We train for 2000 epochs using AdamW with learning rate $3\times 10^{-4}$. In the discovery protocols, eight of the nine behavioral classes are used for supervision, while one class is withheld to simulate novel behavior discovery.

For the containment score, we use boundary mass $0.95$, 2000 Monte Carlo samples, Gaussian KDE bandwidth by Silverman’s rule, and 2D t-SNE perplexity $30$ (other scikit-learn defaults).



\section{Experiments}
\label{sec:results}

\subsection{Experimental Setup}
We evaluate on 4,338 labeled motion snippets from gulls, spanning nine expert-annotated behaviors (\cref{tab:behavior-counts}). Each sample is a 20-step multivariate time series with four channels (3-axis IMU acceleration and GPS speed). We consider two discovery settings: existing novel class discovery and non-existing class discovery (negative control). 

\textbf{Existing novel class discovery.}
We remove one behavioral class from the labeled data and split the remaining samples 50/50. We train the encoder on the first half (containing eight classes). We then embed all samples using this trained encoder and run semi-supervised K-means on the resulting embeddings: the same half used to train the encoder is reused as the labeled set (known classes), while the other half together with all samples from the withheld behavior forms the unlabeled pool. We report withheld-class accuracy and the KDE-based containment score $O_c$ (lower indicates higher novelty).

\textbf{Non-existing class discovery (negative control).}
We keep the same split and clustering protocol, but with no withheld behavior: the unlabeled pool contains only known classes, so the extra discovered cluster should have high containment.

\textbf{Data and code availability.}
Dataset release will follow after documentation is finalized (potentially in a separate dataset publication). Code is available at \href{https://github.com/fkariminejadasl/bird-behavior}{https://github.com/fkariminejadasl/bird-behavior}.

\subsection{Existing Novel Class Discovery}
Across withheld-class experiments, the semi-supervised K-means stage typically isolates the withheld behavior into a distinct cluster, and the containment statistic falls below the novelty threshold ($O_c < 0.3$) for all withheld classes except \texttt{Manouvre} (\cref{tab:existing}). Qualitatively, the t-SNE visualizations show that the withheld class forms a compact region in embedding space that is separated from most known classes. Qualitatively, \cref{fig:tsne_d0} shows a representative withheld-class run, including ground-truth labels and cluster assignments for all data and for the labeled and unlabeled subsets. \cref{fig:all_tsne_lp} summarizes these results across all withheld-class runs.

The \texttt{Manouvre} behavior is the main exception, resulting in a higher containment score $0.496$. A plausible explanation is that \texttt{Manouvre} corresponds to a transition regime between flight modes rather than a distinct state with a compact distribution. Because transitional segments overlap with \texttt{Soar}, and \texttt{Flap} and may be labeled with greater uncertainty, the class distribution in embedding space becomes diffuse and less separable, resulting in higher containment.


\subsection{Non-Existing Class Discovery}
In the negative-control setting, containment scores for the extra discovered cluster remain above 0.3 for all removed-class trials (\cref{tab:nonexisting}). This indicates that the "discovered" cluster is well explained by known-class distributions, and the containment score rejects novelty. The overlap coefficient thus serves as a robust indicator for distinguishing genuine novel classes from spurious discoveries. 

For the negative-control setting, \cref{fig:tsne_r0_d10} shows a representative trial with ground-truth labels and cluster assignments for all data and for the labeled and unlabeled subsets. \cref{fig:all_tsne_lp10} summarizes these results across all negative-control trials.

\subsection{Pipeline Setup}
\label{sec:pipeline_setup}
The online pipeline operates on a sliding window of 100 unlabeled segments. Each segment is a 20-step multivariate sequence in the same format as the training data. We first encode the 100 segments using an encoder trained on the labeled data. We then run semi-supervised K-means with ten clusters, using half of the labeled samples as known-class data and treating the remaining labeled samples together with the 100 unlabeled segments as the unlabeled pool. For each newly formed cluster, we compute the KDE-based containment score $O_c$ and declare a novel behavior when $O_c < 0.3$; otherwise we treat the cluster as a re-partitioning of known behaviors.


\section{Discussion}
\label{sec:discussion}

We considered generalized behavior discovery in short, burst-like motion sequences where (i) labels are scarce and imbalanced, and (ii) the unlabeled pool may contain a behavior absent from the labeled set. Our results suggest that combining label-guided clustering with an explicit novelty statistic provides a practical alternative to purely heuristic cluster inspection. In particular, the KDE + HDR containment score offers an interpretable way to quantify whether a discovered cluster is plausibly explained by any known class distribution.

\textbf{What worked and why.} Across withheld-class experiments, the semi-supervised K-means stage formed coherent candidate groups, and the containment score separated genuine novel behavior clusters (low overlap with all known classes) from the negative-control setting where no novel behavior exists (consistently higher overlaps). This supports the central premise of the paper: discovery pipelines benefit from decoupling \emph{cluster formation} from \emph{novelty decision}. Even when clustering yields a distinct group, the containment score provides a quantitative check against spurious discoveries caused by embedding overlap or cluster fragmentation.

\textbf{Transitional behaviors.} The main failure mode is \texttt{Manouvre}, which attains higher containment and often overlaps with other flight behaviors. A plausible explanation is that \texttt{Manouvre} corresponds to a transition regime between flight modes rather than a distinct state with a compact distribution. Because transitional segments overlap with \texttt{Soar}, and \texttt{Flap}, and may be labeled with greater uncertainty, the class distribution in embedding space becomes more diffuse and less separable, resulting in higher containment.

\textbf{Representation quality is a bottleneck, but the pipeline is encoder-agnostic.} The discovery and containment decision mechanism does not depend on a particular encoder, but its effectiveness depends on embeddings that preserve within-class structure while separating behaviors. In our experiments, off-the-shelf embeddings from large time-series models (MOMENT \cite{moment} and Chronos2 \cite{chronos2}) showed poor separability on this dataset, and additional pretraining of MOMENT on a larger pool $(2.4 \times 10^7)$ of unlabeled data did not resolve this issue. We hypothesize that the mismatch is driven by data regime and objective: our sequences are very short (length 20, padded to 32), whereas these models use patching (patch sizes 8 and 16) that reduces temporal resolution and are trained primarily for forecasting or reconstruction rather than for discovery-oriented invariances.

\textbf{Why standard contrastive pretraining was ineffective.} We also explored MAE-style reconstruction \cite{mae} and InfoNCE-style contrastive learning \cite{infonce}. MAE improved separability relative to foundation-model embeddings but remained insufficient, consistent with reconstruction objectives not explicitly enforcing discriminability. InfoNCE did not yield meaningful representations in our setting, likely due to false negatives induced by small number of classes and strong imbalance, amplified by large batch sizes. Moreover, common time-series augmentations (jitter, scaling, time warping, magnitude warping) appeared to disrupt cues that are discriminative for these behaviors, and similarly reduced accuracy when used for supervised contrastive training.

\textbf{Limitations and future directions.} The novelty threshold (0.3 in our experiments) should be treated as a dataset-dependent operating point; in practice it can be calibrated using negative controls and held-out labeled splits. The containment statistic also inherits limitations of density estimation in a low-dimensional embedding projection, and KDE/HDR estimates can be less stable for small or multi-modal clusters.
A promising direction is burst-aware self-supervised learning that better matches the data regime: variable-length pretraining up to 200 steps, and objectives tailored to discovery rather than forecasting. Positive-only self-distillation methods (for example, DINO-style training \cite{dino}) may reduce false negatives under imbalance. More broadly, hierarchical modeling could encode within-burst motion and then aggregate across bursts with time-gap information, which may better capture behavioral transitions that are difficult to represent from single bursts alone. In addition, we plan to investigate domain-specific augmentations for IMU data, including rotation-based transformations that preserve physical consistency while introducing meaningful variability.

Overall, our findings emphasize that generalized behavior discovery in animal motion data can be made more reliable by pairing semi-supervised clustering with an explicit, interpretable novelty statistic, while leaving room for improved encoders to further strengthen separability and downstream novelty decisions.

\section{Conclusion}
\label{sec:conclusion}
We proposed a semi-supervised pipeline for novel behavior discovery in bird motion data. By combining semi-supervised K-means clustering with KDE-based containment score, our framework enables both recognition of known behaviors and detection of new ones, even under class imbalance and limited annotation. Through experiments on gull motion data, we showed that removing one behavioral class during training allows the model to rediscover it from unlabeled samples. The containment score effectively quantifies this discovery, distinguishing real novel behaviors (low overlap) from spurious ones (high overlap). Our approach provides a general, data-driven method for novel class discovery in ecological time-series data and can be extended to other domains involving partially labeled sequential signals.

\section*{Impact Statement}
This paper introduces a semi-supervised approach for generalized behavior discovery from animal-borne motion time series, aiming to reduce the amount of manual annotation needed in movement ecology studies. The intended impact is to help researchers identify candidate behaviors and quantify novelty in long recordings, potentially improving the speed and reproducibility of behavioral analysis.

A key limitation is that unsupervised or weakly supervised clusters may not correspond one-to-one with biologically meaningful behaviors (for example, they can capture transitions, sensor artifacts, or dataset-specific patterns). We therefore frame discovered clusters as hypotheses that require domain validation and report diagnostic protocols and limitations to reduce the risk of over-interpretation.

\bibliography{paper}
\bibliographystyle{plainnat}

\newpage
\appendix
\onecolumn
\section{Confusion Matrix}
\label{app:confusion-matrix}

The values are reported in \cref{tab:existing}. Since there is only one class to discover, Hungarian matching is not needed. The accuracy for the discovered class is computed as the number of correctly assigned samples divided by the total number of samples in that class. The confusion matrices are computed on $\mathcal{D}_U$.

\begin{figure}[t]
  \centering
   \includegraphics[width=0.98\linewidth]{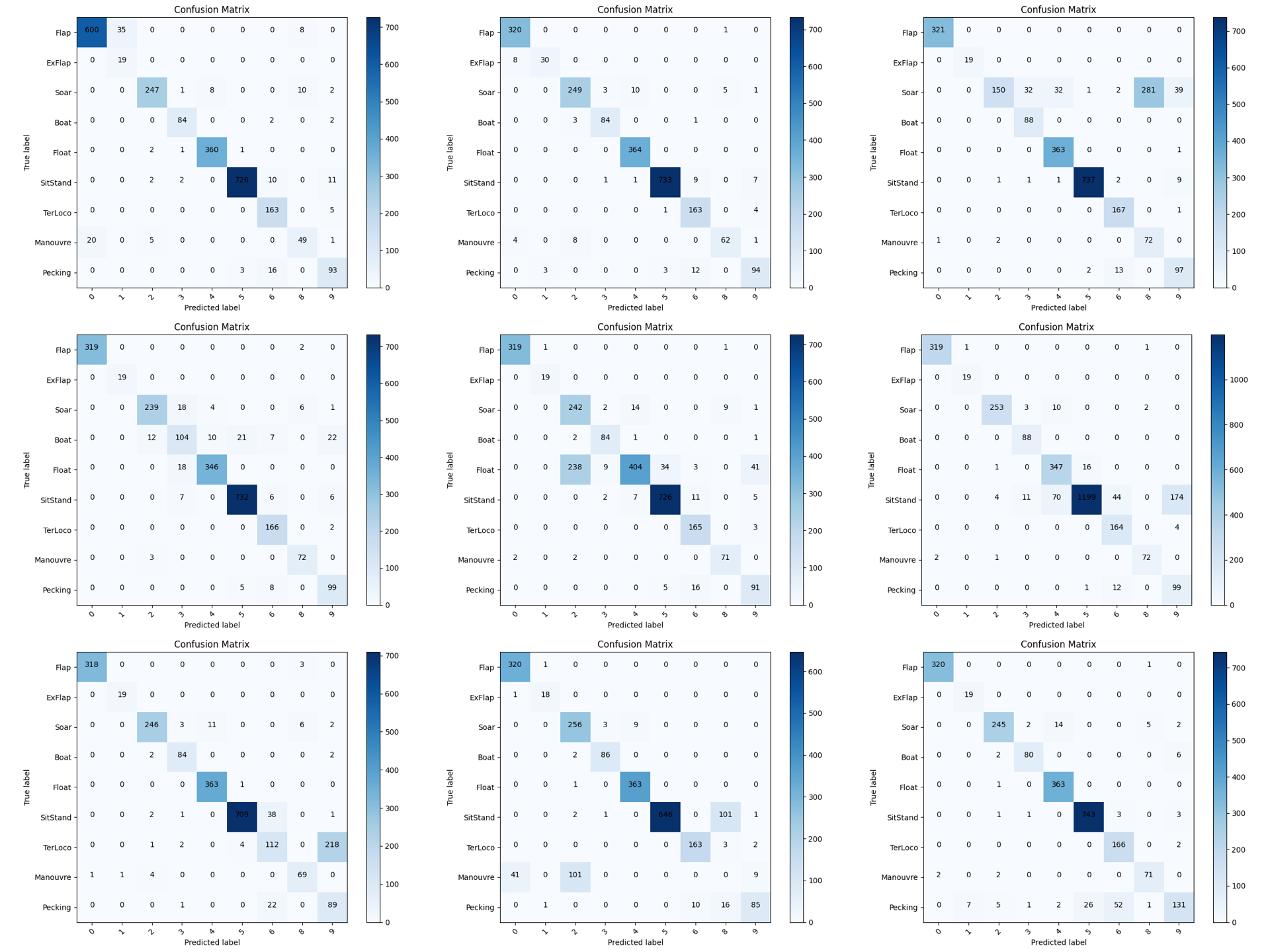}
   \caption{Confusion matrices for existing novel class discovery. The nine withheld-behavior trials are arranged left-to-right and top-to-bottom in the following order: 0:Flap, 1:ExFlap, 2:Soar, 3:Boat, 4:Float, 5:SitStand, 6:TerLoco, 8:Manouvre, 9:Pecking.}
   \label{fig:cm}
\end{figure}

\section{Additional Representation Learning Experiments}
\label{app:rep_learning}

This appendix summarizes representation learning variants we explored to assess whether improved embeddings alone would yield better separation of behavior classes. These experiments are not the main contribution of the paper. They provide context for why we use a lightweight supervised encoder in the main pipeline and delegate novelty decisions to the containment statistic.

\paragraph{Common evaluation protocol.}
For each method, we extract embeddings and visualize class structure using t-SNE. To keep comparisons consistent with the main pipeline, we use the classifier pre-softmax outputs (logits) as embeddings when available. Points are colored by ground-truth behavior labels. All self-supervised pretraining experiments use approximately $2.4\times 10^7$ unlabeled windows.

\begin{figure}[t]
  \centering
  \includegraphics[width=0.98\linewidth]{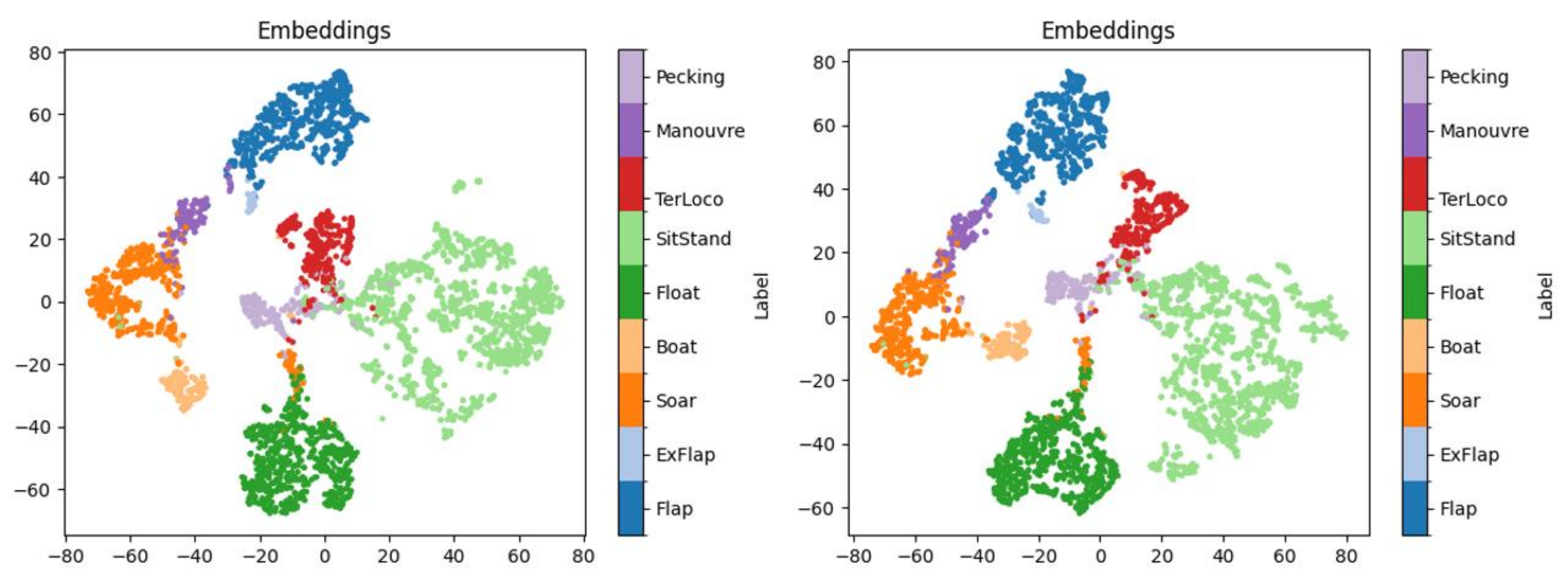}
  \caption{\textbf{Supervised embeddings.} t-SNE of (left) average-pooled features and (right) classifier logits. Colors indicate ground-truth behaviors.}
  \label{fig:125}
\end{figure}

\paragraph{Supervised baseline: logits vs.\ pooled features.}
We compare two embeddings from our supervised temporal model: (i) global average pooled features from the final convolutional block and (ii) classifier logits. As shown in \cref{fig:125}, both yield clear separation, consistent with direct optimization for the labeled taxonomy used during training.

\paragraph{ViT-based MAE-style reconstruction and self-distillation.}
We trained a lightweight ViT encoder-decoder with an MAE-like reconstruction objective and evaluated the resulting embeddings. The model has approximately 10M parameters in total, with roughly 5M in the encoder and 5M in the decoder. Reconstruction improved separation relative to several off-the-shelf embeddings, but remained less discriminative than the supervised baseline, which is expected since reconstruction objectives do not explicitly enforce between-class margins.

\begin{figure}[t]
  \centering
  \includegraphics[width=0.98\linewidth]{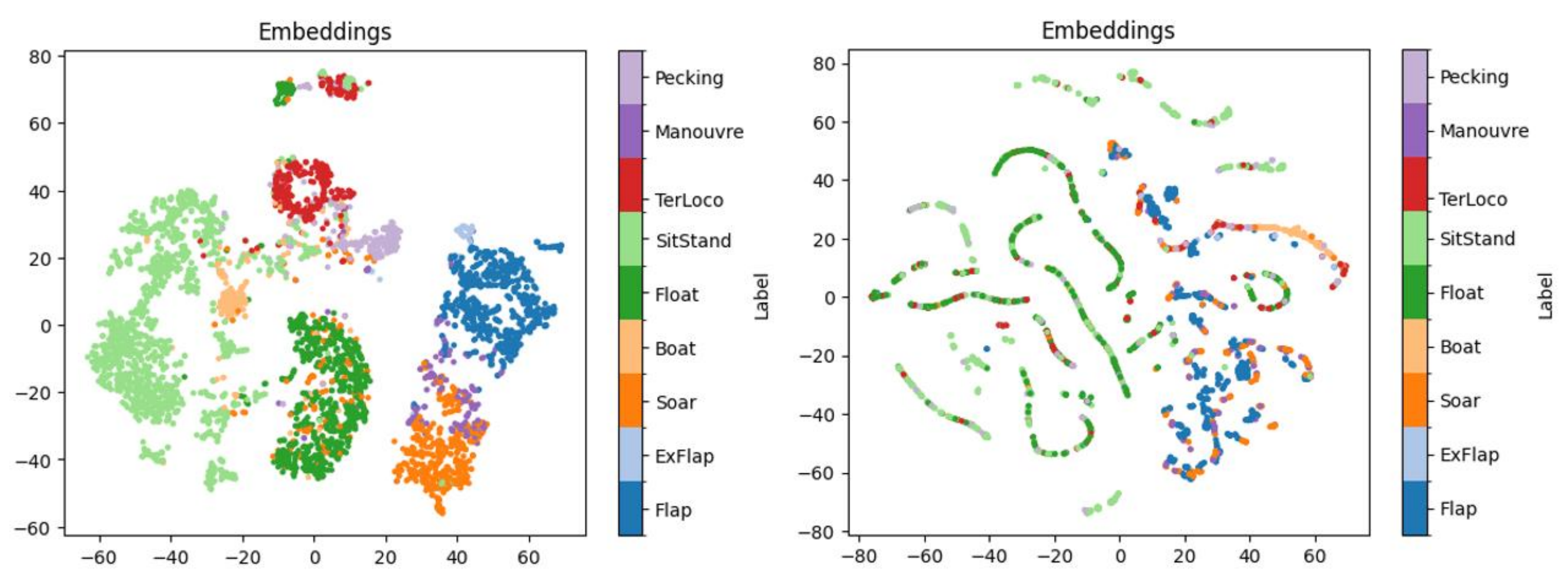}
  \caption{\textbf{Self-supervised ViT variants.} t-SNE of embeddings from (left) MAE-style reconstruction and (right) self-distillation with an InfoNCE loss, initialized from the MAE-trained encoder (trained for one epoch due to rapid degradation). Colors indicate ground-truth behaviors.}
  \label{fig:mae}
\end{figure}

We also evaluated a self-distillation variant that reuses the MAE-trained encoder as initialization and trains it with an InfoNCE-style contrastive loss. In this setting, training rapidly degraded the embedding geometry: after a single epoch, embeddings collapsed into a snake-like manifold with visibly reduced class separation. We therefore report the one-epoch result and did not continue training. A likely contributor is the sensitivity of instance-discrimination to false negatives, which is amplified here by severe class imbalance and short windows, where different behaviors can produce similar local motion patterns and common augmentations can obscure discriminative cues. \cref{fig:mae} provides the qualitative comparison.

\begin{figure}[t]
  \centering
  \includegraphics[width=0.98\linewidth]{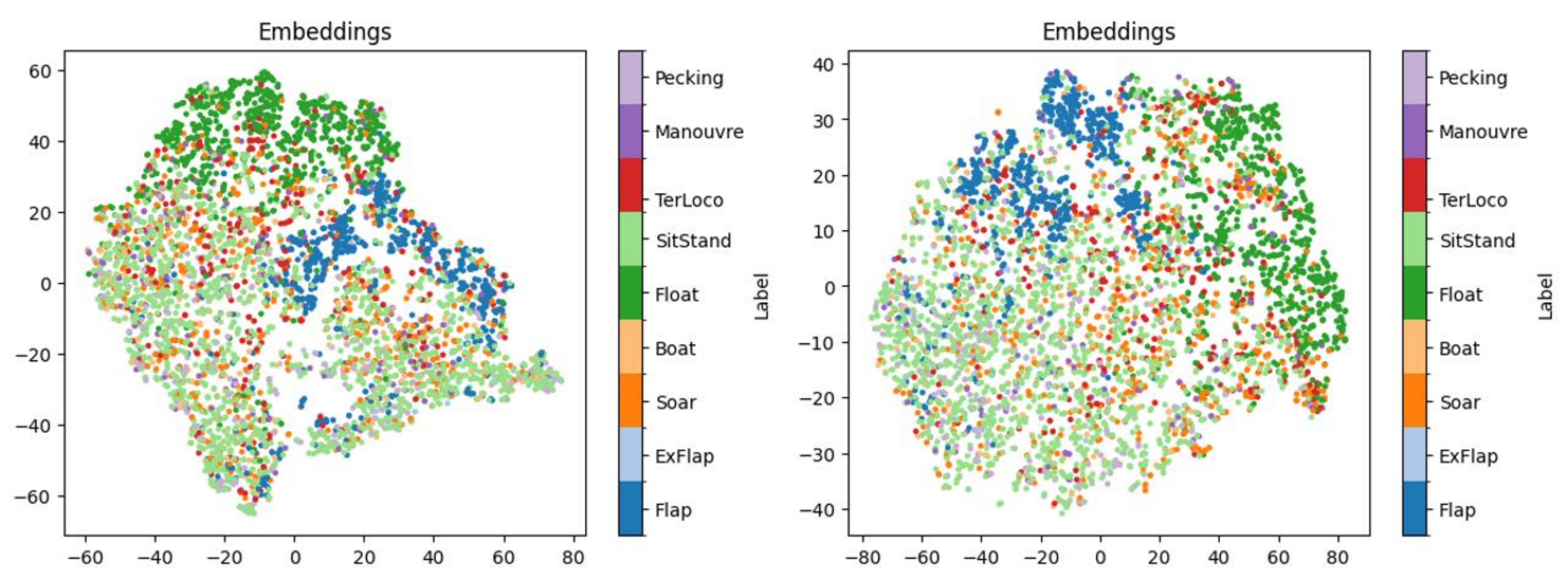}
  \caption{\textbf{Foundation time-series embeddings.} t-SNE of (left) MOMENT released weights and (right) MOMENT finetuned on our unlabeled data. Colors indicate ground-truth behaviors. Despite finetuning, the embeddings show essentially no separability for these short, burst-like windows.}
  \label{fig:moment}
\end{figure}

\paragraph{MOMENT finetuning on gull motion windows.}
We evaluated MOMENT using released weights and after finetuning on our unlabeled pool. We used \texttt{AutonLab/MOMENT-1-small} (approximately 36M parameters) and finetuned all parameters for 20 epochs. Since our windows have length 20, we zero-pad to length 32 to match the model input. MOMENT uses patching with patch size 8, which reduces effective temporal resolution for very short sequences. As shown in \cref{fig:moment}, both released and finetuned embeddings show essentially no class separation on this dataset.

\paragraph{Chronos2 on gull motion windows.}
We extracted embeddings using the released Chronos2 weights (no finetuning) and visualized their structure with t-SNE.

\begin{figure}[t]
  \centering
  \includegraphics[width=0.49\linewidth]{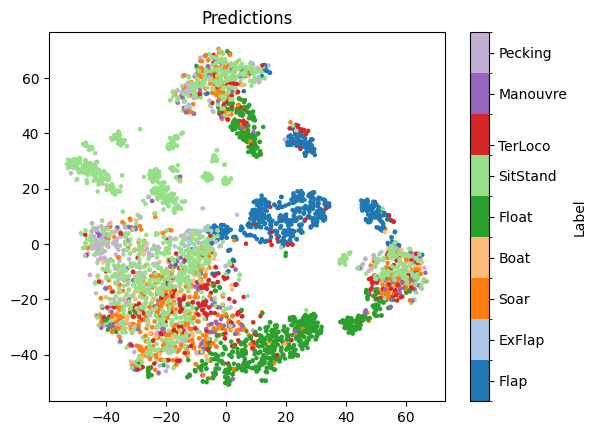}
  \caption{\textbf{Foundation time-series embeddings.} t-SNE of embeddings from Chronos2 (released weights). Colors indicate ground-truth behaviors. The embeddings show limited separability for these short, burst-like windows.}
  \label{fig:chronos2}
\end{figure}

\paragraph{Training details.}
For the ViT encoder-decoder (MAE-like) experiments, we used the following configuration:
\begin{itemize}
  \item Epochs: 150, batch size: 8192
  \item Input: sequence length $g\_len=20$, channels $=4$
  \item Encoder: $embed\_dim=256$, $depth=6$, $num\_heads=8$, $mlp\_ratio=4$ (approximately 5M parameters)
  \item Decoder: $decoder\_embed\_dim=256$, $decoder\_depth=6$, $decoder\_num\_heads=8$ (approximately 5M parameters)
  \item Output channels: $out\_channel=256$, dropout: 0.7
\end{itemize}
For self-distillation with InfoNCE, we used random jitter and scaling augmentations with $\sigma=0.05$ and stopped after one epoch due to qualitative collapse.

For MOMENT, we finetuned \texttt{AutonLab/MOMENT-1-small} for 20 epochs with batch size 256 on 2 GPUs, using input length 32 and zero-padding our length-20 windows.

\paragraph{Takeaway and implications.}
Overall, supervised embeddings provide the most behavior-aligned structure for our discovery pipeline, while self-supervised reconstruction yields partial separation and contrastive self-distillation is unstable in this regime. These observations motivate two directions highlighted in the discussion: (i) positive-only or self-distillation objectives that avoid negative sampling (for example, DINO-style losses) to reduce false-negative pressure under imbalance, and (ii) domain-consistent augmentations, including rotation-based transformations for IMU channels, that preserve physical plausibility while introducing meaningful variability.

\end{document}